\def\year{2020}\relax
\documentclass[letterpaper]{article} 
\usepackage{aaai20}  
\usepackage{times}  
\usepackage{helvet} 
\usepackage{courier}  
\usepackage[hyphens]{url}  
\usepackage{graphicx} 
\usepackage{graphicx}  
\title{LSTM-SAKT: LSTM-Encoded SAKT-like Transformer for Knowledge Tracing}
\author{\\ \Large \textbf{Takashi Oya, Shigeo Morishima}\\ 
Waseda Research Institute for Science and Engineering\\
3-4-1 Okubo, Shinjuku, Tokyo 169-8555, Japan.\\
oya\_takashi@ruri.waseda.jp, shigeo@waseda.jp 
}
 
\begin{document}

\maketitle

\begin{abstract}
This paper introduces the 2nd place solution for the Riiid! Answer Correctness Prediction\footnote{\url{https://www.kaggle.com/c/riiid-test-answer-prediction/}} in Kaggle\footnote{\url{https://www.kaggle.com/}}, the world's largest data science competition website. This competition was held from October 16, 2020, to January 7, 2021, with 3395 teams and 4387 competitors. The main insights and contributions of this paper are as follows. \\
(i) We pointed out existing Transformer-based models are suffering from a problem that the information which their query/key/value can contain is limited. To solve this problem, we proposed a method that uses LSTM to obtain query/key/value and verified its effectiveness.
\\
(ii) We pointed out `inter-container' leakage problem, which happens in datasets where questions are sometimes served together. To solve this problem, we showed special indexing/masking techniques that are useful when using RNN-variants and Transformer.
\\
(iii) We found additional hand-crafted features are effective to overcome the limits of Transformer, which can never consider the samples older than the sequence length.

\end{abstract}

\section{Introduction}
The COVID-19 pandemic from 2020 changed the world overwhelmingly. Specifically, In many countries, it is prohibited or not recommended to communicate face-to-face frequently, including in environments for education. To tackle such situations, some companies have attempted and are attempting to make an AI-tutor, that helps students learn online, by providing questions and lectures that are personalized for each student. Riiid! Labs, a leading solutions provider for AI education, decided to host a competition in Kaggle, the world's largest data science competition website, seeking great systems for AI education. The task of this competition is knowledge tracing, where models need to predict the probability that a certain student can answer a certain question, by using the learning history of the student. As a result, 3395 teams and 4387 competitors participated in this challenge, and the winner's team achieved 0.820 in ROC-AUC score, a major improvement from the current state-of-the-art \cite{saintplus}. In this competition, our team took second place by achieving a score (0.818 in ROC-AUC) that can be comparable to the winner. In this paper, we introduce the 2nd place solution for this challenge, whose contributions and insights can be summarized as these three points, written below.

\noindent
\textbf{(i) LSTM used to obtain query/key/value}
Transformer \cite{transformer} achieved great success in the task of knowledge tracing (e.g. SAKT \cite{sakt}), However, it suffers from a problem that the information which query/key/value can contain is limited. For example, query/key/value of SAKT cannot contain the information of previous samples, so it is hard to consider patterns like `If a certain student solves a vocabulary problem first and then solves a grammar problem, this student tends to make a mistake'. To solve this problem, we leveraged a LSTM model \cite{LSTM} like DKT \cite{DKT}, to obtain query/key/value of SAKT-like Transformer. 

\begin{figure}[t]
\centering
\includegraphics[width=0.4\textwidth]{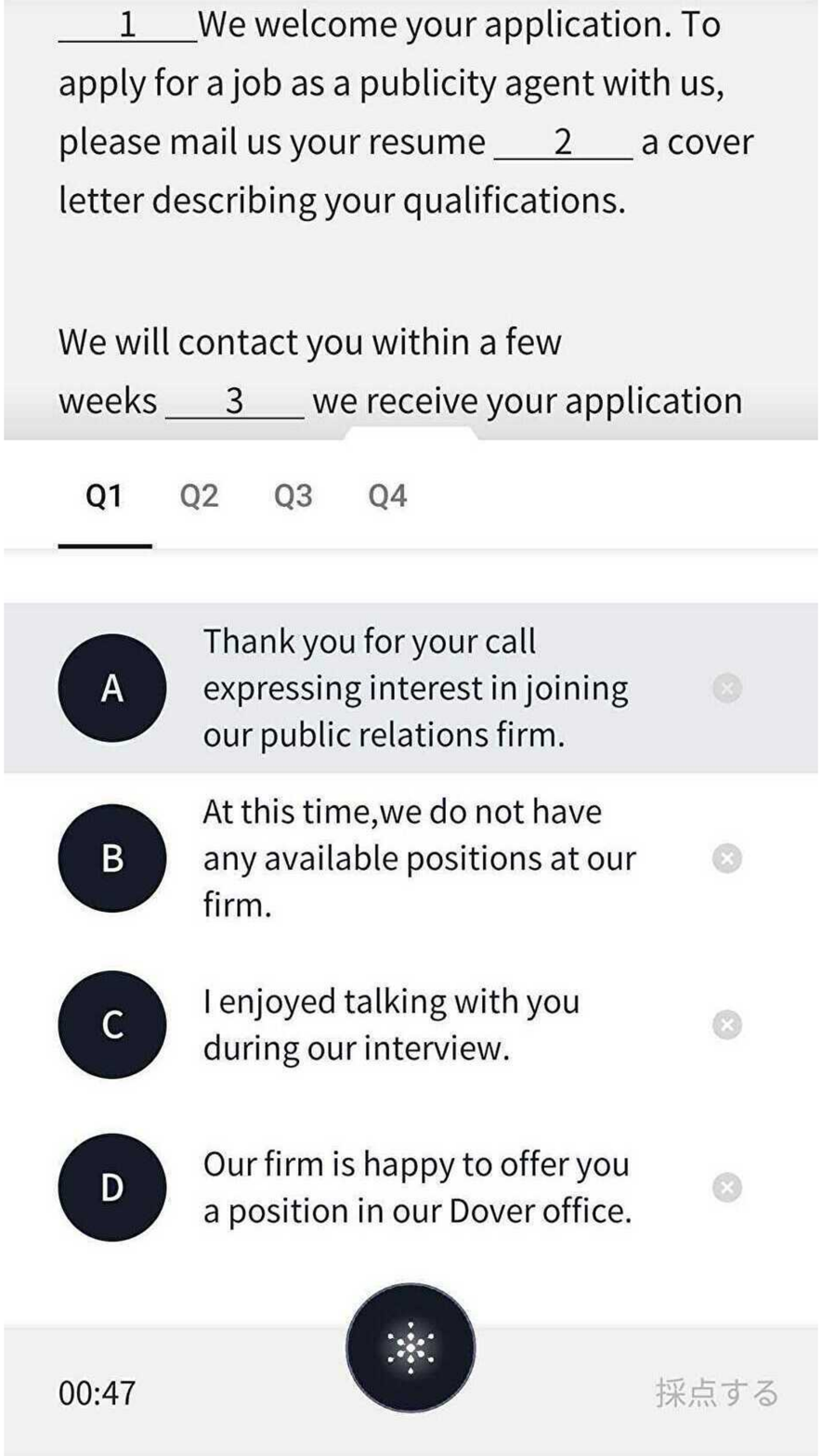} 
\caption{Examples of the problem of SANTA TOEIC, which is a system with AI-Tutor, used for constructing the dataset for the competition. As can be seen, a user has to solve these 4 problems before seeing the answer for any of them.}
\label{toeic}
\end{figure}

\noindent
\textbf{(ii) Special indexing/masking techniques to handle `inter-container' leakage problem}
The datasets used for this competition contains a column 'task\_container\_id', which means some questions are served together. For example, if the task\_container\_id for 4 questions for a certain user is the same, the user has to solve these 4 questions before seeing the answer or explanation for any of them, as shown in Fig. \ref{toeic}. So, when training Transformer or RNN-variants, people can not use some information (e.g. whether the user answered correctly or not) of the samples that share the same task\_container\_id. We call this problem `inter-container' leakage problem. To solve this problem, We used fancy indexing \cite{numpy} and a modified upper triangular mask like the one used in SAINT\cite{saint}.

\noindent
\textbf{(iii) Additional hand-crafted features that contain information which can't be considered by Transformer}
In general, when training Transformer, people set the sequence length and do not consider samples older than the sequence length, without some exceptions (e.g. Transformer-XL \cite{transformerxl}). To overcome this limitation of Transformer, we provided a simple solution, that makes additional hand-crafted features extracted from a user's history. For example, we made a feature `the percentage that the user's answer to the question was correct in all the history of this user'. As expected, such features made an improvement on performance, because such features contain information that Transformer can not capture.

\begin{figure*}[t]
\centering
\includegraphics[width=1\textwidth]{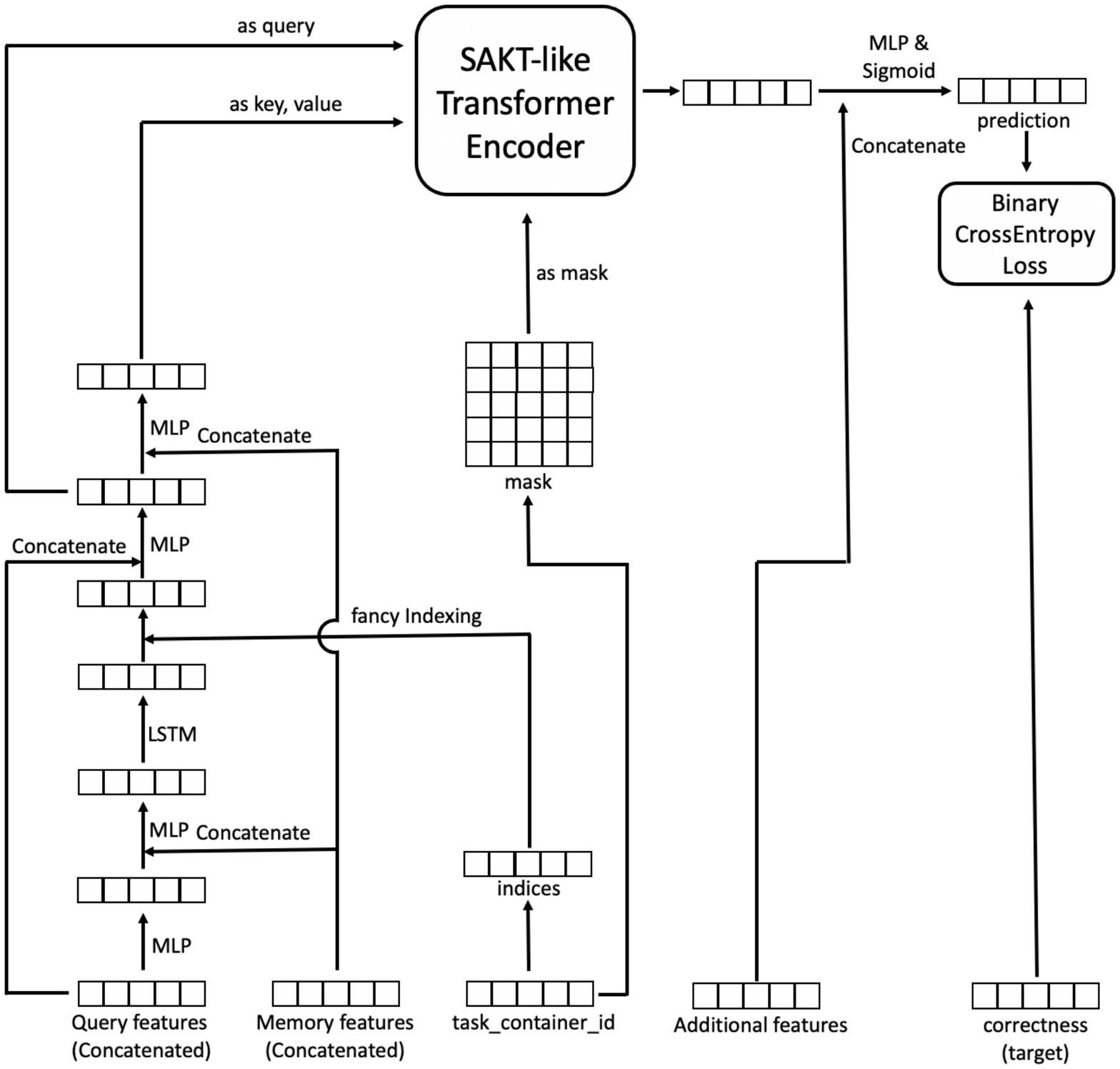} 
\caption{The overall architecture of our model.}
\label{fig2}
\end{figure*}

\begin{figure*}[t]
\centering
\includegraphics[width=1\textwidth]{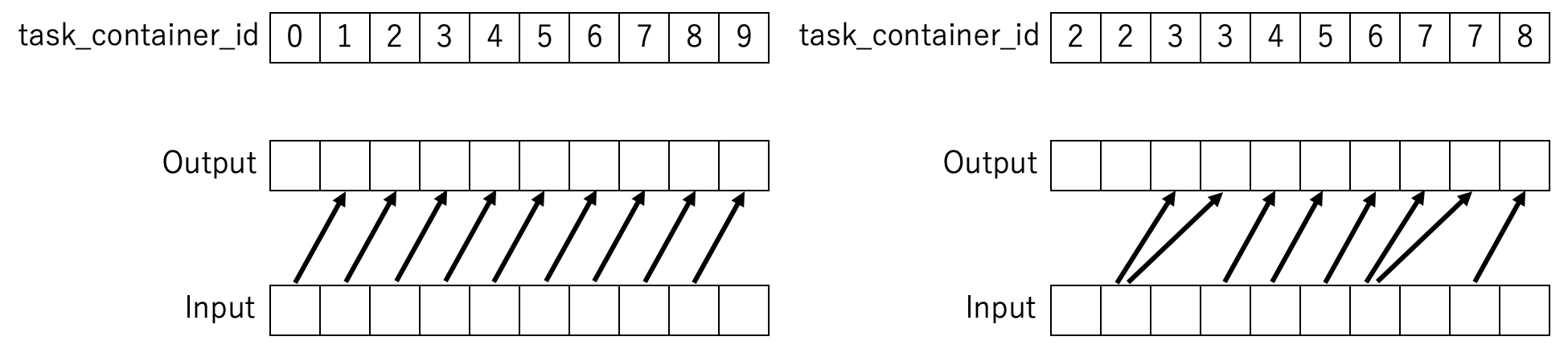} 
\caption{Concept of fancy indexing which is performed after applying LSTM. In the left case, the output is a simple shifted version of the input, as no questions are served together. On the other hand, in the right case, the output is not a simple shifted version of the input, as some questions are served together. }
\label{task}
\end{figure*}

\section{Methods}
In this section, we give a detailed explanation of our model and features, illustrated in Fig. \ref{fig2}. First, we made 3 types of features: query features, memory features, and additional hand-crafted features, whose contents are listed below.
\subsection{Query features}
Below is the list of query features. More details about the column names can be found in the description of the competition\footnote{\url{https://www.kaggle.com/c/riiid-test-answer-prediction/data}} and the description of EdNet datasets\footnote{\url{https://github.com/riiid/ednet}} \cite{ednet}.
\begin{itemize}
  \item \textbf{content\_id}: ID code for the problems. We applied an embedding layer with 512 dimensions to this feature.

  \item \textbf{part}: the section of the TOEIC test. We applied one-hot encoding for this feature.
  \item \textbf{tags}: tag codes for the question. We applied one-hot encoding for this feature.
  \item \textbf{normalized time delta}: the difference between the current timestamp and the previous timestamp. We applied standardization for this feature. 
  \item \textbf{normalized log timestamp}: We calculated $\log(1 + timestamp)$, then applied standardization.
  \item \textbf{correct answer}: the answer to the question. We applied one-hot encoding for this feature.
  \item \textbf{normalized task\_container\_id delta}: the difference between the current task\_container\_id and the previous task\_container\_id, which is then divided by 1000.
  \item \textbf{content\_type\_id delta}: the difference between the current content\_type\_id and the previous content\_type\_id.
  \item \textbf{normalized absolute position}: We used normalized position instead of using positional encoding.
\end{itemize}

\subsection{Memory features}
Below is the list of memory features. It should be noted that memory features are different from query features as memory features are available after students answer the question. 
\begin{itemize}
  \item \textbf{explanation}: whether or not the user saw an explanation of the question after answering the question. We used one-hot encoding for this feature.
  \item \textbf{correctness}: whether the user's answer is correct or not. We used one-hot encoding for this feature.
  \item \textbf{normalized elapsed time}: the amount of time it took the user to answer the question. We applied standardization for this feature.
  \item \textbf{user\_answer}: the user's answer to the question. We applied one-hot encoding for this feature.
\end{itemize}
\subsection{Additional hand-crafted features}
Below are examples of additional hand-crafted features. These features are created to overcome the limitation of Transformer-based models, which can never consider the samples older than the sequence length. We made 90 features as an additional input of our model.
\begin{itemize}
  \item \textbf{correctness features}: the percentage that the user's answer was correct and the number of the problems with which the user interacted in the past.
  \item \textbf{content\_id features}: the percentage that the user's answer to the content\_id was correct and how many times the user interacted with the content\_id in the past. 
  \item \textbf{tag \& part features}: the percentage that the user's answer to each tag/part is correct and how many times the user interacted with each tag/part in the past. 
  \item \textbf{lecture features}: how many lectures the user saw for each part/tag/type\_of in the past.
  \item \textbf{explanation features}: the percentage that the user saw the explanation after they answered the problem correctly/incorrectly and how many times the user saw the explanation.
  \item \textbf{elapsed time features}: the average time it took the user to solve the problem in the past.
  \item \textbf{user\_answer features}: the ratio of the answer (0, 1, 2, 3) the user chose in the past.
  
\end{itemize}

\subsection{LSTM and Fancy Indexing}
After obtaining query features and memory features, we applied MLP \cite{MLP} to query features and then concatenate the output with memory features, and then applied MLP again. Then, we applied LSTM and fancy indexing to avoid the "inter-container" leakage problem. For example, when the task\_container\_id for a specific sample is $[2, 2, 3, 3, 4, 5, 6, 7, 7, 8]$, the indices we use for fancy indexing are $[-1, -1, 1, 1, 3, 4, 5, 6, 6, 8]$, which prohibits the usage of information of the same task\_container\_id. In Fig. \ref{task}, we showed the concept of fancy indexing we did after applying LSTM.

 Here, -1 means the sample of the corresponding position can not use any past information.

\subsection{SAKT-like Transformer and masking}
After applying LSTM, we concatenated the output with query features and applied MLP again, then use it as query for Transformer. After that, we concatenated query for Transformer with memory features and applied MLP again, then use it as key/value for Transformer. To avoid `inter-container' leakage problem, we used a modified upper triangular mask different from SAINT. For example, when the task\_container\_id for a specific sample is $[2, 2, 3, 3, 4, 5, 6, 7, 7, 8]$, the mask is as follows.
$
       \left[1, 1, 1, 1, 1, 1, 1, 1, 1, 1\right], \\
       \left[1, 1, 1, 1, 1, 1, 1, 1, 1, 1\right], \\
       \left[0, 0, 1, 1, 1, 1, 1, 1, 1, 1\right], \\
       \left[0, 0, 1, 1, 1, 1, 1, 1, 1, 1\right], \\
       \left[0, 0, 0, 0, 1, 1, 1, 1, 1, 1\right], \\
       \left[0, 0, 0, 0, 0, 1, 1, 1, 1, 1\right], \\
       \left[1, 0, 0, 0, 0, 0, 1, 1, 1, 1\right], \\
       \left[1, 1, 0, 0, 0, 0, 0, 1, 1, 1\right], \\
       \left[1, 1, 0, 0, 0, 0, 0, 1, 1, 1\right], \\
       \left[1, 1, 1, 1, 0, 0, 0, 0, 0, 1\right]
$

For the hyper-parameters, we used sequence length = 400, dimension = 512, n\_head = 4, dropout = 0.2. We trained our model with lr = 2e-3 for 35 epochs, then trained it with lr = 2e-4 for 1 epoch with the Adam \cite{adam} optimizer. For training, we used 40 Core Xeon CPU, 384GB RAM, V100 (16GB) $\times$ 4.

\section{Conclusion}
In this paper, we described the 2nd place solution for Riiid! Answer Correctness Prediction in Kaggle. Our method achieved a major improvement compared with the current state-of-the-art paper, and performance comparable to the winner of the competition. Additionally, we pointed out the problem of the existing methods and the difficulty of the dataset used for the competition, then proposed the solutions to tackle these problems. We believe the insights obtained in this paper help the experts and the community of the AI-Education area, and hopefully the students who have to learn online due to the COVID-19 pandemic.
\section{Additional Resources}
Additional details are given in the solution writeup in Kaggle\footnote{\url{https://www.kaggle.com/c/riiid-test-answer-prediction/discussion/210113}} and the inference code uploaded at Kaggle notebook\footnote{\url{https://www.kaggle.com/mamasinkgs/public-private-2nd-place-solution}}.

\section{Acknowledgments}
This project is supported by the JST ACCEL (JPMJAC1602),  JST-Mirai Program (JPMJMI19B2) and JSPS KAKENHI (JP19H01129).

\bibliographystyle{aaai}
\bibliography{egbib}

\begin{thebibliography}{}

\bibitem[\protect\citeauthoryear{Choi \bgroup et al\mbox.\egroup
  }{2020a}]{saint}
Choi, Y.; Lee, Y.; Cho, J.; Baek, J.; Kim, B.; Cha, Y.; Shin, D.; Bae, C.; and
  Heo, J.
\newblock 2020a.
\newblock Towards an {Appropriate} {Query}, {Key}, and {Value} {Computation}
  for {Knowledge} {Tracing}.
\newblock {\em arXiv preprint arXiv:2002.07033}.

\bibitem[\protect\citeauthoryear{Choi \bgroup et al\mbox.\egroup
  }{2020b}]{ednet}
Choi, Y.; Lee, Y.; Shin, D.; Cho, J.; Park, S.; Lee, S.; Baek, J.; Bae, C.;
  Kim, B.; and Heo, J.
\newblock 2020b.
\newblock Ednet: A large-scale hierarchical dataset in education.
\newblock In {\em International Conference on Artificial Intelligence in
  Education}.

\bibitem[\protect\citeauthoryear{Dai \bgroup et al\mbox.\egroup
  }{2019}]{transformerxl}
Dai, Z.; Yang, Z.; Yang, Y.; Carbonell, J.; Le, Q.~V.; and Salakhutdinov, R.
\newblock 2019.
\newblock Transformer-{XL}: {Attentive} {Language} {Models} {Beyond} a
  {Fixed}-{Length} {Context}.
\newblock In {\em Association for Computational Linguistics (ACL)}.

\bibitem[\protect\citeauthoryear{Harris \bgroup et al\mbox.\egroup
  }{2020}]{numpy}
Harris, C.~R.; Millman, K.~J.; van~der Walt, S.~J.; Gommers, R.; Virtanen, P.;
  Cournapeau, D.; Wieser, E.; Taylor, J.; Berg, S.; Smith, N.~J.; Kern, R.;
  Picus, M.; Hoyer, S.; van Kerkwijk, M.~H.; Brett, M.; Haldane, A.; del Río,
  J.~F.; Wiebe, M.; Peterson, P.; Gérard-Marchant, P.; Sheppard, K.; Reddy,
  T.; Weckesser, W.; Abbasi, H.; Gohlke, C.; and Oliphant, T.~E.
\newblock 2020.
\newblock Array programming with numpy.
\newblock {\em Nature} 585(7825):357--362.

\bibitem[\protect\citeauthoryear{Hochreiter and Schmidhuber}{1997}]{LSTM}
Hochreiter, S., and Schmidhuber, J.
\newblock 1997.
\newblock Long short-term memory.
\newblock {\em Neural Computation} 9(8):1735--1780.

\bibitem[\protect\citeauthoryear{Kingma and Ba}{2015}]{adam}
Kingma, D.~P., and Ba, J.
\newblock 2015.
\newblock Adam: A method for stochastic optimization.
\newblock In {\em International Conference on Learning Representations (ICLR)}.

\bibitem[\protect\citeauthoryear{Pandey and Karypis}{2019}]{sakt}
Pandey, S., and Karypis, G.
\newblock 2019.
\newblock A self-attentive model for knowledge tracing.
\newblock In {\em International Conference on Educational Data Mining (EDM)}.

\bibitem[\protect\citeauthoryear{Piech \bgroup et al\mbox.\egroup }{2015}]{DKT}
Piech, C.; Spencer, J.; Huang, J.; Ganguli, S.; Sahami, M.; Guibas, L.; and
  Sohl-Dickstein, J.
\newblock 2015.
\newblock Deep {Knowledge} {Tracing}.
\newblock In {\em Neural Information Processing Systems (NeurIPS)}.

\bibitem[\protect\citeauthoryear{Rumelhart, Hinton, and Williams}{1986}]{MLP}
Rumelhart, D.~E.; Hinton, G.~E.; and Williams, R.~J.
\newblock 1986.
\newblock Learning representations by back-propagating errors.
\newblock {\em Nature} 323(6088):533--536.

\bibitem[\protect\citeauthoryear{Shin \bgroup et al\mbox.\egroup
  }{2020}]{saintplus}
Shin, D.; Shim, Y.; Yu, H.; Lee, S.; Kim, B.; and Choi, Y.
\newblock 2020.
\newblock {SAINT}+: {Integrating} {Temporal} {Features} for {EdNet}
  {Correctness} {Prediction}.
\newblock {\em arXiv preprint arXiv:2010.12042}.

\bibitem[\protect\citeauthoryear{Vaswani \bgroup et al\mbox.\egroup
  }{2017}]{transformer}
Vaswani, A.; Shazeer, N.; Parmar, N.; Uszkoreit, J.; Jones, L.; Gomez, A.~N.;
  Kaiser, L.; and Polosukhin, I.
\newblock 2017.
\newblock Attention {Is} {All} {You} {Need}.
\newblock In {\em Neural Information Processing Systems (NeurIPS)}.

\end{thebibliography}

\end{document}